\pdfoutput=1

\documentclass[11pt]{article}

\usepackage{acl}

\usepackage{times}
\usepackage{latexsym}

\usepackage[T1]{fontenc}

\usepackage[utf8]{inputenc}

\usepackage{microtype}

\usepackage{hyperref}       
\usepackage{url}            
\usepackage{booktabs}       
\usepackage{amsfonts}       
\usepackage{nicefrac}       
\usepackage{microtype}      
\usepackage{xcolor}         

\usepackage{multirow}

\usepackage{amsmath}
\usepackage{amssymb}
\usepackage{mathtools}
\usepackage{amsthm}
\usepackage{titlesec}
\usepackage{caption}
\usepackage{subfigure}

\usepackage{CJKutf8}

\definecolor{shadecolor}{rgb}{0.92,0.92,0.92}
\definecolor{Gray}{gray}{0.95}

\usepackage[prependcaption,textsize=tiny]{todonotes}

\newcommand{\datasetname}{\textsc{PersoNet }}
\newcommand{\datasetnamens}{\textsc{PersoNet}}

\title{Personality Understanding of Fictional Characters during Book Reading}

\author{%
  Mo Yu$^{1}$\thanks{\scriptsize{\,\,Authors contributed equally to this paper.}} \quad Jiangnan Li$^{2*}$ \quad Shunyu Yao$^{3}$ \quad Wenjie Pang$^{1}$ \\ 
  \bf \quad Xiaochen Zhou$^{4}$ \quad  Xiao Zhou$^{1}$ \quad  Fandong Meng$^{1}$ \quad  Jie Zhou$^{1}$ \\
  \small{$^1$Pattern Recognition Center, WeChat AI \qquad $^2$Institute of Information Engineering, Chinese Academy of Sciences} \\ \small{$^3$Princeton University \qquad $^4$Syracuse University} \\ 
  \small{\texttt{{moyumyu@global.tencent.com}\quad \texttt{lijiangnan@iie.ac.cn} }}}

\begin{document}
\maketitle
\begin{abstract}

Comprehending characters' personalities is a crucial aspect of story reading. As readers engage with a story, their understanding of a character evolves based on new events and information; and multiple fine-grained aspects of personalities can be perceived.
This leads to a natural problem of \textbf{situated and fine-grained} personality understanding.
The problem has not been studied in the NLP field, primarily due to the lack of appropriate datasets mimicking the process of book reading.
We present the first labeled dataset \datasetname for this problem. Our novel annotation strategy involves annotating user notes from online reading apps as a proxy for the original books. Experiments and human studies indicate that our dataset construction is both efficient and accurate; and our task heavily relies on long-term context to achieve accurate predictions for both machines and humans.\footnote{\scriptsize{Available at \url{https://github.com/Gorov/personet_acl23}.}}

\end{abstract}

\section{Introduction}
\label{sec:intro}

Lively fictional characters with distinct personalities are the first drive of the plotline developments.
The authors shape the characters with various personality types, which distinguish a character from others and explain the motivations and behaviors of the characters.
As a reverse process, the readers grasp the characters' personalities during reading a story, which helps to understand the logics of a plot and predict its future developments. 

The NLP community has also recognized the values of personality understanding; and conducted studies~\cite{bamman2013learning,flekova2015personality,sang2022mbti} along this direction.
In the problem definition of 
the existing tasks, the input is an entire book.
By construction, they ask for the general impression of character personalities.
Also for this reason, they only focus on coarse-grained personality types, \emph{e.g.}, the four coarse MBTI types~\cite{myers1988myers}.

\begin{figure}[ht]
\centering
\vspace{-0.1in}
\includegraphics[width=0.45\textwidth]
{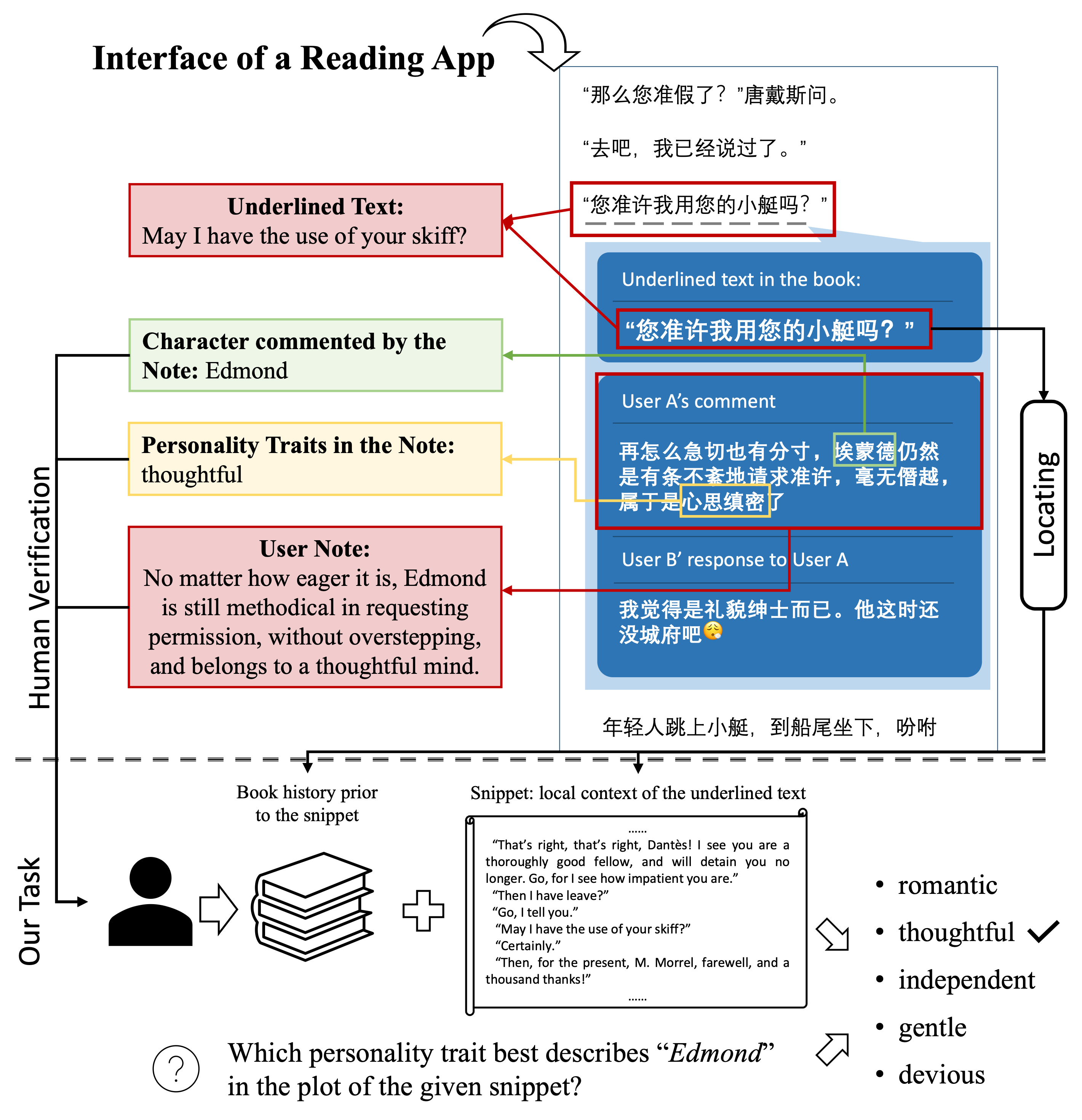}
\vspace{-0.15in}
\caption{\small{Illustration of an reading app interface enabling our dataset construction (user IDs masked for privacy); and an example of our task, \emph{situated personality prediction}.}}
\vspace{-0.2in}
\label{fig:weread_app}
\end{figure}

To make a personality prediction task more practical and useful,
we consider two important aspects of character understanding in real life that have not been studied in the context of machine reading.
First, we aim at predicting \textbf{fine-grained} personality types, with an exhaustive vocabulary of personality traits as the targets.
Second and more importantly, we study the continuous-process nature of story reading --- As people read, they form dynamic impressions of the characters and plots.
We name this process \textbf{situated} comprehension.
Specific to personality understanding, a character may have multi-faced personalities. In a certain point of the story, the character's behaviors can reflect one of them when faced the situation and  events at the time.
Human readers have the ability to use their knowledge of what has happened so far (\emph{i.e.,} the \textbf{history}) to understand the character in the current situation.
We hence propose to study \textbf{situated personality prediction}, which differs from the static prediction problem studied before.

While the aforementioned two problems are practical and common in real life,
they create new challenges in dataset creation, especically the latter.
To accurately mimic the human reading process, annotators would need to read entire books, which is not practical due to the significant time required.

We overcome this annotation difficulty and create a large-scale dataset for personality prediction in the reading processes. 
To achieve this goal, we propose a novel annotation strategy that utilizes publicly available book notes.
Recent online reading apps such as Kindle, Douban, iReader\footnote{\scriptsize{\url{https://book.douban.com},  \url{https://www.ireader.com.cn}.}} allow users to take notes while reading a book (an example shown in Figure~\ref{fig:weread_app}).
As users read, they can add notes at the current reading position. These notes are linked to specific text in the book, which is highlighted with a dotted underline, referred to as \textbf{underlined texts} throughout this paper. This mechanism ensures that the notes accurately reflect the thoughts the user had while reading the surrounding text of the underlined text.

Based on this resource, 
we propose our strategy of annotating \emph{user notes as a delegate of the book reading process}.
Specific to our task of personality prediction,
this corresponds to (1) identifying if a user note discusses the personality trait of a character; and (2) associating the trait label to the context at the note location.
We take user notes that contain at least a character name and a personality trait word, and ask human annotators to confirm if the trait is a modifier of the character in the note text (\emph{i.e.}, the user note mentions that the character has the trait).
The verified notes serve as nature labels of character personalities reflected by the surroundings of the underlined texts.
By using this approach, we collect labeled data that only requires annotators to read short notes, without the need for knowledge about the books themselves.

With our new strategy, we create our situated personality prediction dataset, \datasetnamens, that contains $\sim$32K instances from 33 books in the classic literature domain.
We prove that our annotation strategy is \emph{efficient} as each worker only requires a median of 23.7s to finish one sample.
The whole annotation process costed in total \$2,400 and 471.8 hours (distributed to 20 working days by 11 annotators).
It is also \emph{accurate} evidenced by both the over 88\% inter-annotator agreement.
In addition, we make the dataset bilingual in both English and Chinese with automatic book sentence alignment and manual character alignment.

We conduct experiments with our dataset in two folds.
First, we develop various improvement over the standard pre-trained models, including enabling the models to use different types of long contexts, equipping the models with oracle history trait information, and task-oriented unsupervised training. 
Second, we conduct extensive human studies with people who have read the books (\emph{i.e.}, with the knowledge of the book history) and not.
Our results show that (1) our task is challenging as humans with knowledge of book history can achieve more than 70\% accuracy, compared to the best model accuracy of $\sim$45\%;
(2) our task heavily requires the long context modeling, as introducing characters' history information significantly improves the model accuracy; and humans without the book history can only perform on par with models.

We make the following contributions:

\noindent$\bullet$ A dataset, \datasetnamens, that is the first benchmark of \emph{situated reading comprehension} and of \emph{fine-grained personality prediction} on books. 
We prove that our dataset is a valid assessment to long context understanding for both machines and humans without significant shortcuts.

\noindent$\bullet$ A novel dataset creation approach for book comprehension problems based on user notes, which is efficient and accurate. 

\noindent$\bullet$ 
Task-oriented unsupervised training and character history enhancement methods that improve on our task, providing insights to future work.

\section{Related Work}
\label{sec:related}

Story book understanding has been recognized as a challenging and rewarding direction~\cite{piper2021narrative,sang2022survey}.
Many evaluation benchmarks on various narrative understanding tasks have been developed, such as plot structure analysis~\cite{saldias2020exploring,papalampidi2019movie}, question answering~\cite{richardson2013mctest,kovcisky2018narrativeqa,xu2022fantastic}, summarization~\cite{ladhak2020exploring,kryscinski2021booksum,chen2021summscreen}, character identification and relationship extraction~\cite{elson2010extracting,elsner2012character,elangovan2015you,iyyer2016feuding,chaturvedi2016modeling,kim2019frowning,sang2022tvshowguess}.

All of the prior work takes the entire long story as input to a model for predictions.
None of them considers the \emph{situated} reading process like ours.

Existing strategies of dataset construction over \textbf{long stories}
fall into the following categories:
$\bullet$A straightforward way is \textbf{to have labelers read the entire stories}. Because of the huge efforts, it only works for short stories for young children~\cite{xu2022fantastic} or simpler tasks like coref~\cite{bamman2019annotated}, NER~\cite{bamman2020litbank} and quotation attribution~\cite{vishnubhotla2022project}.
$\bullet$\textbf{Using the book summaries as proxy} of the original stories, \emph{e.g.}, the creation of book-level question answering task~\cite{kovcisky2018narrativeqa}.
    The created data usually only covers abstract and major events in the book, as shown in~\cite{mou2021narrative}. Thus the types of comprehension skills that can be assessed with this strategy are limited.
$\bullet$\textbf{Exploiting Web resources created by fans or experts}. 
\citet{flekova2015personality} used fans' rated MBTI types to create a classification task for book characters; \citet{ladhak2020exploring,kryscinski2021booksum} created a book chapter summarization task based on summaries on the English learning websites; and \citet{relic22} created a book retrieval task based on quotes in literature reviews. 
    The drawback of this strategy is that the tasks can be supported are limited by the available resources.
\noindent$\bullet$\textbf{Automatically created cloze tests} is a traditional strategy. With specifically designed techniques, the clozes can be made resolvable only with global context, \emph{e.g.},~\cite{raecompressive2019,sang2022tvshowguess,yu2022few}. The problem of this method is that the created datasets usually have unclear assessment goals.

The limitations of these strategies make them insufficient to create datasets for our task of situated personality understanding.

\section{Problem Definition}
\label{sec:problem}

Our \datasetname is the first task on situated prediction of characters' personality traits in book contexts.
That is, we aim to predict the traits reflected by a local snippet of book, given all the previous book content as the background (Figure~\ref{fig:weread_app}). 

Formally, we consider a local book snippet ${\mathcal{S}^{(i)}}=\{s_{k^{(i)}_1}, s_{k^{(i)}_2}, ..., s_{k^{(i)}_{J}}\}$. Each $s_{k^{(i)}_j}$ is a sentence from the book, with $k^{(i)}_j$ the absolute position of the sentence in the book.
Each ${\mathcal{S}}$ in our task depicts a character's personality. Therefore, it is associated with a pair of $(c, t)$, where $c$ is a character name or alias and $t$ is the personality trait of $c$ that reflected by $\mathcal{S}$. Note that different pairs may share a same snippet.
Our task is then to predict:
\begin{equation}
\small
\begin{aligned}
    P(y = t\vert c, \mathcal{S}^{(i)}, \mathcal{H}^{(i)}=s_{1:k_1^{(i)}}),\label{eq:problem}
\end{aligned}
\end{equation}
where $s_{1:k_1^{(i)}}$ refers to all the sentences before $\mathcal{S}^{(i)}$ in the book.
We split the books into training, dev and test sets, so that the evaluation characters are unseen during training.
For evaluation, we adopt a multi-choice setting. For each instance, we sampled 4 negative candidates, two from the top-20 most frequent traits and the rest from the whole list.
Combining the negative choices with $t$, we have a candidate set $\mathcal{T}$. Our data thus form a tuple $(\mathcal{S}, \mathcal{H}=s_{1:t_k^{(i)}}, c, t, \mathcal{T})$.

\section{Our \datasetname Dataset}
\label{sec:dataset}

\subsection{Data Source}
\label{ssec:source}
\medskip
\noindent\textbf{List of Personality Traits }
Following previous work~\cite{shuster2019engaging}, we use the list of 818 English personality traits from the MIT Ideonomy project.\footnote{\scriptsize{\url{http://ideonomy.mit.edu/essays/traits.html}.}}
We translate the traits into Chinese with Youdao dictionary,\footnote{\scriptsize{\url{https://cidian.youdao.com/}.}} then ask human annotators to select all the translated meanings that depict personality in Chinese.
There are 499 English traits and 565 Chinese traits left that are bilingually aligned.
 
\medskip
\noindent\textbf{Books and Notes }
We collect 100 public books available in the Gutenberg project.
For each book, we find its Chinese-translated versions that we have licenses of usage; then collect all their public user notes
from the Internet.
We kept notes that (1) contain any traits, (2) contain any person names\footnote{\scriptsize{We use Spacy (\texttt{zh\_core\_web\_lg}) for NER.}} and (3) with lengths less than 100 words (relatively shorter notes can improve human annotation efficiency).
We filtered out the books with less than 100 notes left, leaving 33 books and 194 of their Chinese translations.
These books have 110,114 notes that contain 140,268 traits in total.

\medskip
\noindent\textbf{Note Clustering }
It is common for multiple users to comment on the same part of a book, discussing the same character. When these users express similar opinions about a character, it leads to duplication.
To remove this duplication for data annotating, we group the notes according to their positions, defined as the center token offset of its associated snippet $\mathcal{S}^{(i)}$ (\emph{i.e.}, its underlined text).
Notes with distances smaller than 100 tokens are grouped, leading to 27,678 
note clusters.
We take the unique traits within each cluster for human labeling, which corresponds to 113,026 samples as defined in Section~\ref{sec:problem}. 
The notes are anonymized for human annotation.

\medskip
\noindent\textbf{Extension of the Snippets }
The lengths of underlined texts can vary significantly, which means they may not always provide a representative context for reflecting a character's personality, particularly when the texts are very short. 
We address this issue by extending each $\mathcal{S}^{(i)}$ from the underlined text to a window of 480 tokens. This window is generally large enough to encompass a scene and ensures that the context relevant to the user note is included.
The reason for choosing this window size is that it is typically longer than one page displayed by the reading app (as shown in Figure~\ref{fig:weread_app}) --- users often write notes on the same page while reading the context, rather than flipping through previous or subsequent pages.\footnote{\scriptsize{Therefore, if the context is not covered by the window, it suggests that the note should not be taken on that page.}}

\subsection{Dataset Construction}
\label{ssec:construction}
Our dataset construction consists of two major steps: (1) human annotation of user notes; (2) projection of labeled data from Chinese to English.
In addition, we show that (3) our data construction strategy enables to build an accurate note classifier for automatic weakly-supervised data labeling.

\medskip
\noindent\textbf{Step 1: Human Annotation }
This step requires the annotators to read each user note, and determine if it discusses the personality of a character.
We present the annotators with notes that contain at least one trait word in our vocabulary in Section~\ref{ssec:source}. 
The note is paired with the \emph{underlined book content}, which is optional to read, if they think the note itself is ambiguous.
The annotators are then asked to (1) judge if the note is indeed about a certain character's trait; then (2) marked the target character name with the trait from the note.

The first step takes most of the human efforts. We wrote concrete guidelines (Figure~\ref{fig:guideline} in Appendix~\ref{app:interface}) for the decision making process. The annotators are citizens in China who have received at least high school education (which, in the Chinese education system, covers most of the general knowledge about classic literature). Therefore it is more convenient for them to work in Chinese; and Figure~\ref{fig:guideline} lists both the original guidelines in Chinese and their English translations.

Our annotation interface (with English translations) is shown in Appendix~\ref{app:interface}. Once the annotators confirm that the given trait word describes some characters, they are required to annotate the character name by dragging from the note text. If not, the character name will be left empty.

\medskip
\noindent\textbf{Step 2: Bilingual Projection }
The human annotation step has created a personality prediction dataset in Chinese.
Next we project the data to English. 
Since the same English book may have multiple translated books in Chinese, their labeled data scattered.
By projecting the labeled data to English books, the book version is unified and the annotations become dense.

According to Section~\ref{sec:problem}, to create an English version of our dataset, we only need to project the traits $t$, the characters $c$ and the snippets (positions) $\mathcal{S}$. The trait $t$ is already from a bilingual vocabulary, so we only need to focus on the latter two.

\noindent$\bullet$\textbf{Book Alignment}
The projection of $\mathcal{S}$ is equivalent to finding each labeled instance's sentence positions in the English book, which is essentially a sentence alignment problem. 
Specifically, we sentencize the books firstly with Spacy; then utilize the \textit{vecalign}~\cite{thompson-koehn-2019-vecalign}
toolkit to achieve sentence alignments among books. We represent each sentence with the default number (10) of its consecutive sentences, and employ the multilingual sentence embedding \textit{LASER}\footnote{\scriptsize{\url{https://github.com/facebookresearch/LASER}.}} 
to embed the sentences. After that, \textit{vecalign} performans sentence alignments with dynamic programming based on the embeddings.

With bilingual sentence alignment, the position of each labeled instance can be mapped to the corresponding position in the English book, \emph{i.e.}, $\mathcal{S}_{\text{en}} = \{a(s) \vert \forall s \in \mathcal{S}\}$, where $a(s)$ refers to aligned position of the Chinese sentence $s$ in the English book.
For most of the $\mathcal{S}$ in our dataset, we can find consecutive $\mathcal{S}_{\text{en}}$ as the aligned results.
There are a few instances mapped to empty. We excluded these cases in our English dataset. There are also a few instances mapped to inconsecutive English sentences, sometimes in a wide range. For this situation, we take the median position of the mapped English sentences and include the consecutive context in a window as the projection. 

\noindent$\bullet$\textbf{Character Name Projection}
We manually project the list of 377 frequent (appear $>$10 times in our labeled data) character names to English.
We askeds two annotators to find the English names of these characters;
and resolved all the inconsistency after they complete their own annotation jobs.

\medskip
\noindent\textbf{Step 3: Weakly-Supervised Data }
Our method reduces the problem of annotation over books to annotation over notes.
This makes it possible to build a note classifier for automatic data augmentation.

We collect another 65,521 notes from the same book collection that consists of at least one trait word and one person name.
By pairing traits with names within the same notes, we create 154,030 examples.
Then we train a binary \emph{roberta-wwm-ext}~\cite{cui-etal-2020-revisiting} classifier over our human-labeled data to determine if the note discusses the character's trait, \emph{i.e.}, the same task in human annotation but without the need of marking target characters.
For each human annotated note, if the note is recognized as describing a trait of a character, it is used as a positive example.
For those labeled as irrelevant to character traits, \emph{i.e.}, no characters are annotated, we denote them as negative examples.
Cross-validation on the human-labeled data shows that our classifier is accurate: 91.1\% and 90.2\% on the dev and test set.
Applying our classifier to these unlabeled examples, we recognize 31,346 examples as describing characters' traits.

\subsection{Quality of the Annotated Data}
\label{ssec:quality_evaluation}
This section proves the accuracy of our data construction method via human study.

\medskip
\noindent\textbf{Correctness of Book Notes}
First of all, we need to prove that the user notes are indeed an accurate delegate of books.
That is, when a note mentions a personality of a character, whether it is highly consistent to what the book content reflects.

This study requires annotators who have read the books to make the correct judgement.
We selected four books with two annotators who have read and are familiar with them.
Each annotator labeled two books.
We sampled in total 431 notes from these books.
The annotators are required to judge if the note is accurate about the character or not. We present the corresponding underlined content along with the note, so that the annotators can identify which part the note is commenting.
The results in Table~\ref{tab:study1} show that 89.1\% of the notes are accurate understanding of the books.
There are 9.7\% \emph{ambiguous} examples, meaning the annotated traits are implied by the current place of the books, but might be falsified later, \emph{e.g.}, the authors may intend to mislead the readers to create surprisal or tension.
These ambiguous labels give valid data for our problem of \emph{dynamic personality prediction}, according to our description at the beginning of Section~\ref{sec:intro} and Eq. (\ref{eq:problem}).

\begin{table}
  \begin{minipage}[b]{0.23\textwidth}
\small
\centering
\begin{tabular}{c||c} 
\toprule
correct & 89.1 \\
ambiguous & 9.7 \\
incorrect & 1.2 \\
\bottomrule
\end{tabular}
\caption{\small{Notes (\%) that consistently reflect the character personalities in the stories.}}
    \label{tab:study1}
\end{minipage}
\hspace{0.06in}
  \begin{minipage}[b]{0.23\textwidth}
\small
\centering
\begin{tabular}{c||c} 
\toprule
perfect match & 187 \\
high overlap & 7 \\
low overlap & 1 \\
no match & 5 \\
\bottomrule
\end{tabular}
\caption{\small{Human study: quality of bilingual alignment.}}
\label{tab:study2}
\end{minipage}
\end{table}

\medskip
\noindent\textbf{Accuracy of Human Labels}
Next, we proved that our annotation process leads to accurate human labels.
This accuracy is verified in two ways. 
First, we compute the inter-annotator agreement, with a duplicated set of 3,000 notes during annotation.
88.67\% of the duplicated samples receive consistent labels.
The Cohen's Kappa~\cite{cohen1960coefficient} is 0.849, which indicates nearly perfect agreement~\cite{viera2005understanding}.
Second, as shown in the Step 3 in Section~\ref{ssec:construction}, a fairly accurate note classifier can be trained on our human-labeled data (91.1\% and 90.2\% accuracy on dev and test).

Both tests confirm the accuracy of our annotation strategy.
Considering the relevance of the book notes (Table~\ref{tab:study1}), this gives an estimation of overall accuracy around 87.6$\sim$89.1\%. The two endpoints are computed with inter-annotator agreement and classifier accuracy accordingly.
It confirms that our dataset is overall accurate.

Table~\ref{tab:ambiguous_cases} in Appendix~\ref{app:difficult_cases} gives some difficult examples that created disagreements.
There are two major sources of difficulties: (1) the trait word has multiple meanings in Chinese and the usage does not represent the sense of the trait; (2) a trait word is used to recall the general impression or history behavior of a character in an implicit way.

\begin{table}[t!]
\setlength{\belowcaptionskip}{-.5\baselineskip}
\setlength{\abovecaptionskip}{0.5\baselineskip}
    \small
    \centering
    \renewcommand{\arraystretch}{1} 
    \scalebox{0.93}{
    \begin{tabular}{lcc||cc} 
        \toprule
        \bf Set & & & \multicolumn{2}{c}{\bf \bf \#Instance}  \\ 
        & \bf \#Books & \bf \#Chars & \bf English & \bf Chinese  \\
        \midrule
        Train  & \multirow{2}{*}{17}& \multirow{2}{*}{148} &18,190 & 18,273   \\
        \quad Weakly Sup  && &26,244  &26,331\\
        Development  &6 &54 & 3,745 &3,751  \\
        Test  &10 &72 & 3,624 & 3,647\\
        \midrule
        Total  & 33 & 274 & 51,803  & 52,002 \\
        \bottomrule
    \end{tabular}
    }
    
    \caption{\small{Data statistics of our PersoNet dataset, including the number of unique books, characters and the numbers of instances in English and Chinese datasets. The weakly supervised data is used for training only.}}
    \label{tab:stats}
    \vspace{-0.1in}
\end{table}

\begin{figure*}
\centering
\subfigure[\scriptsize{Dant\`{e}s (\emph{i.e.}, Count Monte Cristo)}]{\includegraphics[width=0.3\textwidth]
{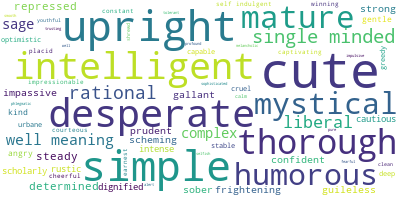}\label{cloud_a}}\
\subfigure[\scriptsize{Albert (\emph{The Count of Monte Cristo})}]{\includegraphics[width=0.3\textwidth]
{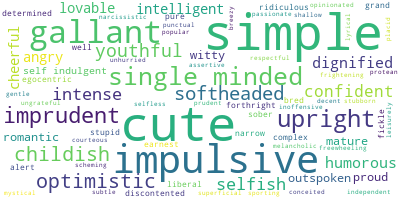}\label{cloud_b}}\
\subfigure[\scriptsize{Plots of sentiments of traits along time}]{\includegraphics[width=0.34\textwidth]{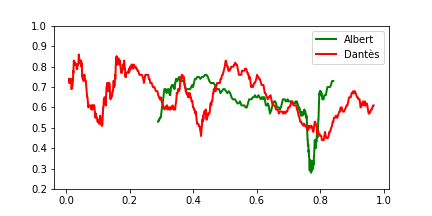}\label{plot_a}}
\vspace{-0.15in}
\caption{\small{Word clouds and plots of sentiments of traits along time for the characters.}}
\vspace{-0.1in}
\centering
\label{fig:word_clouds}
\end{figure*}

\medskip
\noindent\textbf{Accuracy of Cross-Lingual Alignment }
Finally, we evaluate the quality of the bilingual alignment.
We randomly sampled 200 labeled instances for human study.
We present to the annotators the snippet $\mathcal{S}$ of each instance in the Chinese book and their aligned sentences from the English books.
The human annotators were asked to rate the alignments into four grades: \emph{perfect}/\emph{high overlap}/\emph{low overlap}/\emph{no match}, \emph{i.e.}, all/$>$50\%/$<=$50\%/none of the Chinese sentences have their translations in the paired English sentences.
Table~\ref{tab:study2} show that $>$97\% of the cases fall into the \emph{perfect} and \emph{high overlap} categories.
When taking texts from the median position of the sentences for model inputs, these categories both can make accurate projections of annotations to the English books.

\subsection{Data Statistics and Visualization}

\paragraph{Data Statistics}
Table~\ref{tab:stats} shows the statistics of our \datasetnamens. We give the full list of books in Appendix~\ref{app:book_list}.
We can also see that our dataset contains a wide range of book characters.
In the annotated training set, approximately 41\% of the notes are about positive traits, 36\% are about negative traits, and 23\% are about neutral traits. This distribution reveals a slight bias, which can be attributed to the fact that users are more inclined to write notes when they have strong sentiments or opinions about a character.

\medskip
\noindent\textbf{Visualization of Our Dataset }
Figure~\ref{fig:word_clouds} visualizes the major traits and the polarity of traits over time for two of the most popular characters. It can be found that the major traits match readers' common impressions; and the trends well align with the common feelings of readers during reading.
This further confirms the quality of our data.

Detailed explanations of the figures and more examples can be found in Appendix~\ref{app:visualization} and Figure~\ref{fig:sentiments_full}.

\section{Models for Persona Prediction}
\label{sec:methods}

We design models based on two different types of pre-trained models, BERT~\cite{devlin2018bert} and Longformer~\cite{beltagy2020longformer}.
We use the latter model to investigate the strength of models that are pre-trained to handle long contexts.

\subsection{Input to the Reader Models}\label{ssec:model_input}
Our data instance consists of a tuple $(\mathcal{S}, \mathcal{H}, c, t, \mathcal{T})$. Here $\mathcal{S}$ is a book snippet that expresses a personality trait $t$ of a character $c$.  
$\mathcal{H}$ is the previous history of $\mathcal{S}$ in the book. 
$\mathcal{T}$ is a set of candidate traits with $t$ as an element.
The task is to rank $t$ to the top within $\mathcal{T}$ given $(\mathcal{S}, \mathcal{H})$ and $c$.
We represent the input $(\mathcal{S}, \mathcal{H}, c)$ with the following format options:

\noindent$\bullet$\textbf{No history} represents the input as $x=[c \text{ \small{[SEP]} } \mathcal{S}]$, \emph{i.e.}, does not use the history $\mathcal{H}$.

\noindent$\bullet$\textbf{Extended history:} $x=[c \text{ \small{[SEP]} } \mathcal{S} \text{ \small{[SEP]} } \mathcal{H}_{\text{prev}}]$, where $\mathcal{H}_{\text{prev}} \subset \mathcal{H}$ includes sentences that are adjacent to $\mathcal{S}$, truncated by models' length limited.

\noindent$\bullet$\textbf{Character history:} $x=[c \text{ \small{[SEP]} } \mathcal{S} \text{ \small{[SEP]} } \mathcal{S}_{c}]$. $\mathcal{S}_{c} \subset \mathcal{H}$ includes snippets to the left of $\mathcal{S}$ that contains the character $c$ in our dataset.

\subsection{Model Architectures}
Our methods compute the score of an input $x$ having a trait $t$, based on the siamese model.

\paragraph{Text Encoding}
Firstly, we use a pre-trained LM (PLM, either BERT or Longformer) to encode $x$ and $t$ to the embedding space. The encoded contextualized embeddings of input and output are denoted as $\mathbf{X} = \text{PLM}(x) \in \mathbb{R}^{l_x \times d}$, where $l_x$ is the length of $x$ and $d$ is the size of hidden states;
and $\mathbf{T} = \text{PLM}(t) \in \mathbb{R}^{l_t \times d}$, with $l_t$ the length of $t$.

\paragraph{Baseline Siamese Model}
As our baseline models, we compute a weighted sum over $\mathbf{X}$ to get a vector representation of the input.
Specifically, we use a linear model to compute the attention score over each token of $x$:
\vspace{-0.05in}
\begin{equation}
\small
\begin{aligned}
    A &= \textrm{Attention}(\mathbf{H}), \, \alpha = \textrm{Softmax}(A). \nonumber
\end{aligned}
\vspace{-0.1in}
\end{equation}
The attention $\alpha_x$ is then used to summarize the hidden states $\mathbf{X}$ a vector $\mathbf{x} = \mathbf{X}^T \alpha$.

The sequence of a trait $t$ is usually short (\emph{e.g.}, a single word's BPE tokenization). Therefore we simply take the average $\mathbf{t} = \text{mean}(\mathbf{T})$.
The model makes prediction with $t = \arg\max_{t \in \mathcal{T}} $$<$$\mathbf{x},\mathbf{t}$$>$.

\paragraph{Contextualization with History}
When the input $x$ contains the extended or character history as defined in Section~\ref{ssec:model_input}, we need to separate the information of the history from the current context.
We maintain a mask $H \in \mathbb{R}^{l_x \times 1}$, such that $H[j]=1$ if the $j$-th word belongs to the appended history and 0 otherwise. Two attention vectors are computed for the current snippet and the history:
\vspace{-0.07in}
\begin{equation}
\small
\begin{aligned}
    \alpha_s = \textrm{Softmax}(A \odot (1-H)), \, \alpha_h = \textrm{Softmax}(A \odot H). \nonumber
\end{aligned}
\vspace{-0.1in}
\end{equation}
The corresponding summaried vectors are $\mathbf{s} = \mathbf{X}^T \alpha_s$ and $\mathbf{h} = \mathbf{X}^T \alpha_h$.
The prediction function is then modified with a gating function $\sigma(\mathbf{s})$ added:
\begin{equation}
\small
\begin{aligned}
    t = \arg\max_{t \in \mathcal{T}} \sigma(\mathbf{s})<\mathbf{s},\mathbf{t}> + (1-\sigma(\mathbf{s}))<\mathbf{h},\mathbf{t}>.\label{eq:model_with_history}
\end{aligned}
\end{equation}

\subsection{Unsupervised Training}
Finally, we propose an unsupervised training task to improve personality prediction. 
The unsupervised task is used to pre-train the classifiers, before they are fine-tuned on our labeled data.
The task mimics the problem definition in Section~\ref{sec:problem} and constructs tuples of $(\mathcal{S}, t)$.
We first extract sentences that contain trait words.
If a sentence $s_j$ contains a trait $t$, we keep a local window of it as the book snippet, with the sentence itself removed.
That is, 
${\mathcal{S}^{(i)}}=\{s_{j-w}, \cdots, s_{j-1},  s_{j+1}, \cdots, s_{j+w}\}$.
Intuitively, since $\mathcal{S}$ provides the context of $s_j$, it is informative for inferring the appearance of the trait described in $s_j$. Therefore this unsupervised task helps to find narrative clues of traits thus can help to better pre-train the encoders.

The method has the limitation of not being character-specific, 
hence not compatible with our character-history-based models. We leave it to future work.

\section{Experiments}
\label{sec:exp}

\subsection{Experimental Settings}
We use \emph{bert-base-uncased} and \emph{longformer-base-4096} as backbones for English experiments; and \emph{Roberta-wwm-ext} for the Chinese experiments.

\medskip
\noindent\textbf{Hyperparameters } For our siamese models with and without history, the most important hyperparameter is the lengths of $\mathcal{S}$ and $\mathcal{H}$. We set the maximal length of $\mathcal{S}$ to 480 tokens for most of the models. For models with history we set the maximum of $\vert \mathcal{S}\vert+\vert \mathcal{H}\vert$=1,600.
To show the better performance of our usage of history, we also compare with Longformer with a maximum $\vert\mathcal{S}\vert$=2K tokens (the best a single A100 GPU can handle).

The batch size is 40 for BERT-based models; and 8 for Longformer-based models with gradient accumulation every 5 batches. Each epoch of BERT and Longformer models takes $\sim$7 and $\sim$40 minutes respectively on a single A100 GPU.  We set the learning rate to 2$e^{-5}$. We conduct early-stopping on the dev set; and run 5 times to compute the average and stand derivation for all the methods.

\medskip
\noindent\textbf{Additional Baselines }
Besides the models in Section~\ref{sec:methods}, we further compared with the follows:
\noindent$\bullet$\textbf{Models with Oracle Traits in History}, which uses the character's history traits in replace of the history texts. For each instance, we take its target character $c$'s other instances prior to it, and concatenate their groundtruth traits as a sequence to replace $\mathcal{H}$ in the model of Eq. (\ref{eq:model_with_history}).
$\bullet$\textbf{Char-Majority}, which always predicts the most frequent trait for a character.
This is used to show the diversity of traits for the same character (\emph{i.e.}, necessity of situated prediction).
\noindent$\bullet$\textbf{GPT-davinci} (text-davinci-003), the few-shot instruct-GPT~\cite{ouyang2022training}.
\noindent$\bullet$\textbf{ChatGPT}, which conduct zero-shot prediction on our task thus can take longer inputs. We test $\vert \mathcal{S}\vert$=480 and 1.6K as in our experiments with trained models.
\noindent$\bullet$\textbf{Humans:} we present the same format of our instances with maximal $\vert \mathcal{S}\vert$=480 to humans to get their performance.

Furthermore, we added LoRA~\cite{lora2022iclr} fine-tuning of the \textbf{LLaMA}~\cite{meta2023llama} and \textbf{WeLM}~\cite{su2022welm} on our \datasetname as additional baselines. The fine-tuning of large language models and the usage of ChatGPT reflect the latest state-of-the-arts in concurrence with our work.

\subsection{Overall Results}

Our main results are shown in Table~\ref{tab:overall_performance_en}. First, all the three models without the usages of history achieve similar results. The Longformer with a 2K window does not give better performance, showing that simply increasing the length of input without including useful history information is not helpful for our task.
Second, our model with character history achieves the best results. Replacing the character history with extended history slightly reduces the dev performance but lead to significant test performance drop (according to the standard derivation).
Among all the supervised-only methods, this model is the only on that maintains consistent dev and test accuracy.
Third, our unsupervised training significantly improve the accuracy for all the models.

Fourth, the oracle history traits improve the supervised accuracy with a large margin. Yet for Longformer, adding character history and unsupervised training makes the gap smaller.
Finally, the best human performance with knowledge of story history greatly outperforms all the models with and without oracle information with 20$\sim$23\%, showing the challenges and great potential of our \datasetnamens.
These results highlight the importance of incorporating history information in solving our task; and reveal that characters exhibit dynamic personalities that evolve over time, thus solely relying on history traits (even oracle) is limited. 

The two methods based on large language models, namely GPT-davinci and ChatGPT, performed worse than the models trained on our dataset. 
This indicates that our task is still a challenge for these general-purpose models. Moreover, although ChatGPT performed better than GPT-davinci, it was not better overall to use the longer context length of 1.6K as compared to using shorter contexts. 
This suggests that ChatGPT may not have been trained to effectively utilize long context in our situated reading setting.

\begin{table}[t!]
\setlength{\belowcaptionskip}{-.5\baselineskip}
\setlength{\abovecaptionskip}{0.5\baselineskip}
    \small
    \centering
    \renewcommand{\arraystretch}{1} 
    \begin{tabular}{lccc} 
        \toprule
        \bf System &\bf Len & \multicolumn{2}{c}{\bf Accuracy}\\ 
        && \bf Dev & \bf Test\\
        \midrule
        Random & -- &20.00 & 20.00   \\
        Frequent Traits & -- &14.10 & 12.75   \\
        \midrule
        BERT (no-hist) & 480 & 45.01\tiny{$\pm$0.64} &  42.96\tiny{$\pm$1.07} \\
        \quad + unsup & 480&{46.18}\tiny{$\pm$0.49} & 44.93\tiny{$\pm$1.01} \\
        Longformer (no-hist) & 480  & 45.02\tiny{$\pm$0.45} & 42.75\tiny{$\pm$0.97} \\
        Longformer (no-hist) &2K & 45.00\tiny{$\pm$0.44} & 42.42\tiny{$\pm$0.39} \\
        Char-Hist-Longformer & {1.6K}& {45.50}\tiny{$\pm$0.54} & {45.33}\tiny{$\pm$1.11} \\
        \quad + unsup &1.6K&\textbf{46.39}\tiny{$\pm$0.63} & \textbf{45.85}\tiny{$\pm$0.72} \\
        \quad w/ extend-hist& 1.6K&45.46\tiny{$\pm$0.67} & 43.44\tiny{$\pm$0.72}\\
        \quad\quad + unsup &1.6K&45.93\tiny{$\pm$0.52} & 44.54\tiny{$\pm$1.49} \\
        \midrule
        \multicolumn{4}{c}{\underline{\emph{w/ Oracle Information}}}\\
        BERT + hist traits & 480&  50.15\tiny{$\pm$1.01} & 50.02\tiny{$\pm$1.03}  \\
        Longformer + hist traits & 2K& 48.66\tiny{$\pm$0.96} & 48.11\tiny{$\pm$1.21}  \\
        Char-Majority &-- & 16.10 & 17.25\\
        \midrule
        GPT-davinci 5-shot$^{*}$ &480& 34.88 & 31.51 \\
        ChatGPT 0-shot$^{*}$ &480& 33.72 & 42.47\\
        ChatGPT 0-shot$^{*}$ &1.6K& 36.05 & 36.99 \\
        LLaMA + LoRA-sft$^{*}$ &1.6K& 47.67 & 49.32 \\
        Human w/o history$^{*}$ &480& 44.19 &40.54 \\
        Human w/ history$^{*}$ &480& 69.77 & 65.75\\
        \bottomrule
    \end{tabular}
    \caption{\small{Overall
    performance (\%) on our \datasetnamens-en task.
    (*) Results were conducted on a subset of the dataset.}}
    \label{tab:overall_performance_en}
\end{table}

\paragraph{Chinese Task Performance}
Table~\ref{tab:overall_performance_zh} shows results on the Chinese version of \datasetnamens.
The results are in general higher than those in the English setting for two reasons:
(1) during annotation we have the semantic space of traits in Chinese, so their English translations may not be the most commonly used words.
(2) the user notes tend to reuse words in the books, so there is higher change that some traits explicitly appear in Chinese books.

\paragraph{Performance of Fine-Tuned LLMs}
To fine-tune the LLMs, we adopt the same setup in the ChatGPT experiments, where the same prompts serve as inputs and the ground truth answers are used as outputs.
The optimization focuses on minimizing perplexity concerning the outputs.
Regarding hyperparameter tuning, we specifically adjust the rank $r$, weight $\alpha$, and number of training epochs. 
For model selection, we rely on the accuracy on the development subset utilized in our human study, which sets $r=8$, $\alpha=1$ and 10 training epochs.

The results in Table~\ref{tab:overall_performance_en} and \ref{tab:overall_performance_zh} show that the fine-tuned LLM achieves slightly better results compared to our proposed baselines.
However, it still significantly lags behind human performance by a considerable margin.
Interestingly, unlike the other models and humans, the fine-tuned LLM perform better on the testing subset compared to the development one. Our hypothesis is that the testing book \emph{Notre-Dame de Paris} is more popular on the Internet, thus may be more sufficiently trained during the pre-training stages of LLaMA and WeLM.

The LLM fine-tuning results can be potentially improved by employing a contrastive training approach similar to our proposed models. We leave this to future study.

\begin{table}[t!]
\setlength{\belowcaptionskip}{-.5\baselineskip}
\setlength{\abovecaptionskip}{0.5\baselineskip}
    \small
    \centering
    \renewcommand{\arraystretch}{1}
        \begin{tabular}{lcc} 
        \toprule
        \bf System & {\bf Dev}  & {\bf Test}  \\ 
        \midrule
        BERT Reader  & 49.72 &  48.70 \\
        Multi-Row BERT Reader  & 50.25 &  49.25 \\
        \midrule
        BERT w/ Trait-History & 53.29 &51.25 \\
        \midrule
        GPT-davinci 5-shot$^{*}$ & 33.72 &32.78 \\
        ChatGPT 0-shot$^{*}$ & 34.88 & 41.89 \\
        WeLM + LoRA-sft$^{*}$ & 51.16 & 54.05\\
        Human w/ history$^{*}$ & 73.26 &68.92 \\
        \bottomrule
    \end{tabular}
    \caption{\small{5-choice accuracy (\%) on our \datasetnamens-zh task.}}
    \label{tab:overall_performance_zh}
\end{table}

\subsection{Human Study}
\label{ssec:exp_human_study}
We conduct human study to understand the challenges of our task.
We sampled instances from the two books that have most instances from the development and testing sets; and asked human annotators (who are co-authors of the paper but have not seen the labeled data before) to complete our multi-choice task.
There are two types of annotators: Type-I who have not read the books before (\emph{human w/o history}); and Type-II who have read the books (\emph{human w/ history}).

We have annotated in total 160 samples. Each sample is guaranteed to be annotated by two humans, one with history and one without history.

\paragraph{Ratio of Ambiguous Instances}
Sometimes an event in a book can depict multiple aspects of personality.
When the sampled negative choices share similarity to these personality traits, it leads to ambiguous cases with more than one correct answers.
To investigate these cases, we require the Type-II annotators to mark the instances that they believe have ambiguous labels.\footnote{\scriptsize{Because these people have memory of the books, they can accurately distinguish the ambiguous cases from those can be disambiguated by the history.}}
There are 41 ambiguous samples recognized, \emph{i.e.}, $\sim$25\% of the cases have more than one correct answers.
This indicates an \textbf{$\sim$87.2\% approximated upperbound} accuracy of our task, if we consider each ambiguous instance has two choices that are correct.

In the future, we can leverage our note clusters to mitigate this ambiguity by ensuring that negative candidates do not appear in the cluster from which the snippet originates.

\paragraph{Main Findings}
The knowledge of book history is not only important to models, but also to humans.
Table~\ref{tab:human_study_results} compares humans performance with and without history.
There is an $\sim$25\% performance gap.
Furthermore, human performance without history is only comparable to the best model performance (selected according to dev accuracy, which performs 47.18\% and 47.21\% on the full dev and test data).
These results confirm that our task raises the core challenge of long context understanding.

Detailed results show that the Type-I annotators labeled $\sim$35\% of cases that they believe unsolvable because of their lacking of the book history.
After verification by Type-II annotators, there are 37 cases left for close examination.
It reveals that the history information is critical for these cases for two major reasons: (1) there are multiple possible answers given the snippets but with the knowledge of the characters' history behavior the incorrect traits can be resolved (17 of 37); (2) the plots in the snippets cannot be understood and linked to any personality without book history (11 of 37).
There is a third difficult category (9 of 37), where reasoning is required to draw connections, \emph{e.g.}, consequence or analogy between the current snippet and a character's previously demonstrated personality.
Examples of these categories can be found in Table~\ref{tab:cases_need_history} in Appendix~\ref{app:examples_history}.

\begin{table}[t!]
    \small
    \centering
    \renewcommand{\arraystretch}{1}
    \begin{tabular}{lcc} 
        \toprule
        \bf System & \multicolumn{2}{c}{\bf Data} \\
        &{\bf All}  & {\bf Unamb}  \\ 
        \midrule
        Best model & 48.75 & 49.58 \\
        \midrule
        GPT-davinci 5-shot & 33.33 & 38.46\\
        ChatGPT zero-shot& 37.74 & 41.03 \\
        LLaMA + LoRA-sft & 48.43 & 52.63 \\
        \midrule
        Human w/o history & 42.50 & 50.42 \\
        Human w/ history& 67.92 & 73.50 \\
        \bottomrule
    \end{tabular}
    \caption{\small{Comparison of performance among models and humans. The \emph{Unamb}iguous subset is annotated by annotators who have read the books.}}
    \label{tab:human_study_results}
\end{table}

\subsection{Analysis}

\begin{figure}
\centering
\vspace{-0.3in}
\includegraphics[width=0.45\textwidth]
{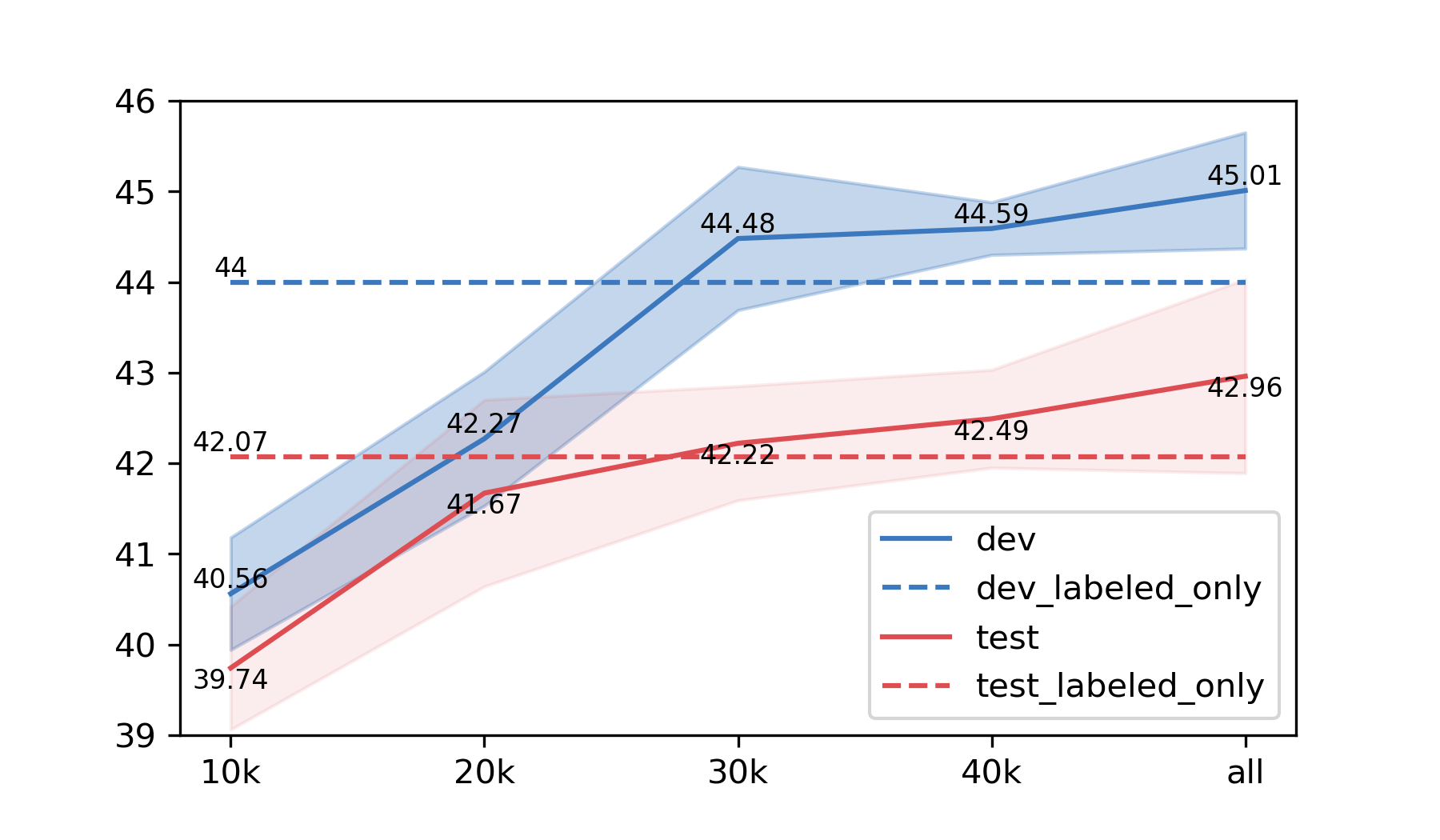}
\vspace{-0.15in}
\caption{\small{Learning curves with varying sizes of training data.}}
\centering
\label{fig:learning_curve}
\end{figure}

\paragraph{Learning Curve}
Figure~\ref{fig:learning_curve} plots the learning curve of our \datasetname task.
The curves shows that the size of our dataset is large enough as the curves become flat after the point of 30K.
More importantly, the results justify the accuracy of our data construction strategy.
As adding weak supervision (all) significantly outperforms training with only human-labeled data (dotted lines).
\paragraph{Difficult Trait Types}
We examine the traits that 
appear more than 20 times in the dev set. The most difficult types include \emph{Confident} (0.00\%), \emph{Mature} (5.56\%), \emph{Liberal} (7.41\%), \emph{Humorous} (7.69\%), \emph{Impressionable} (8.82\%), \emph{Gentle} (9.09\%), \emph{Optimistic} (10.81\%), \emph{Rational} (11.36\%), \emph{Imprudent} (14.29\%) and \emph{Insincere} (16.00\%).
It can be found that most of the difficulty types are abstract, which are usually not explicit depicted in the books but require reasoning from characters' behaviors.



\section{Conclusion}
We propose a dataset \datasetname for the new problem of situated personality understanding of book characters.
We overcome the difficulty in dataset construction with a new strategy of annotating the user notes as a proxy for the original books.
Our dataset constuction method maintains both efficiency and accuracy.
Experiments show that the task raised challenges of long-text understanding for both humans and machines.

\section*{Limitations}
Our propose annotation strategy can be applied to labeling other MRC problems, no matter situated comprehension ones or not.
However, when generalizing to other problems other than personality prediction we studied here, the accuracy of the user notes may vary with the difficulty of tasks.
Additional human verification on the correctness of notes like in our Section~\ref{ssec:quality_evaluation} need to be conducted.

Our unsupervised training technique does not support
the Longformer reader with character history (Char-Hist Longformer) yet.
Therefore, the improvement from unsupervised training for our this model is smaller.

While Longformer is common in benchmarking for long story understanding tasks. There are other families of models~\cite{raecompressive2019,izacard2020leveraging,ainslie2020etc,xiong2021nystromformer,pang2022long} handling long text encoding. We leave the comparison with these models to future work.

\paragraph{Potential Risks} Like the other work that based on the similar set of  books~\cite{bamman2019annotated,bamman2020litbank,vishnubhotla2022project,relic22}, the classic literature may be limited by the time of writing, thus raise fairness considerations.
However, please note that our dataset construction strategy is not limited to these books, but can work with any books on the reading apps to create a sampled book set without such biases.
The main reason we stick with the current list of books is for reproducibility since they are publicly available.

\bibliography{anthology,custom}
\bibliographystyle{acl_natbib}

\clearpage
\appendix

\section{Annotation Guidelines and Interface}
\label{app:interface}

\begin{figure*}
\setlength{\belowcaptionskip}{-.5\baselineskip}
\centering
\includegraphics[width=\textwidth]{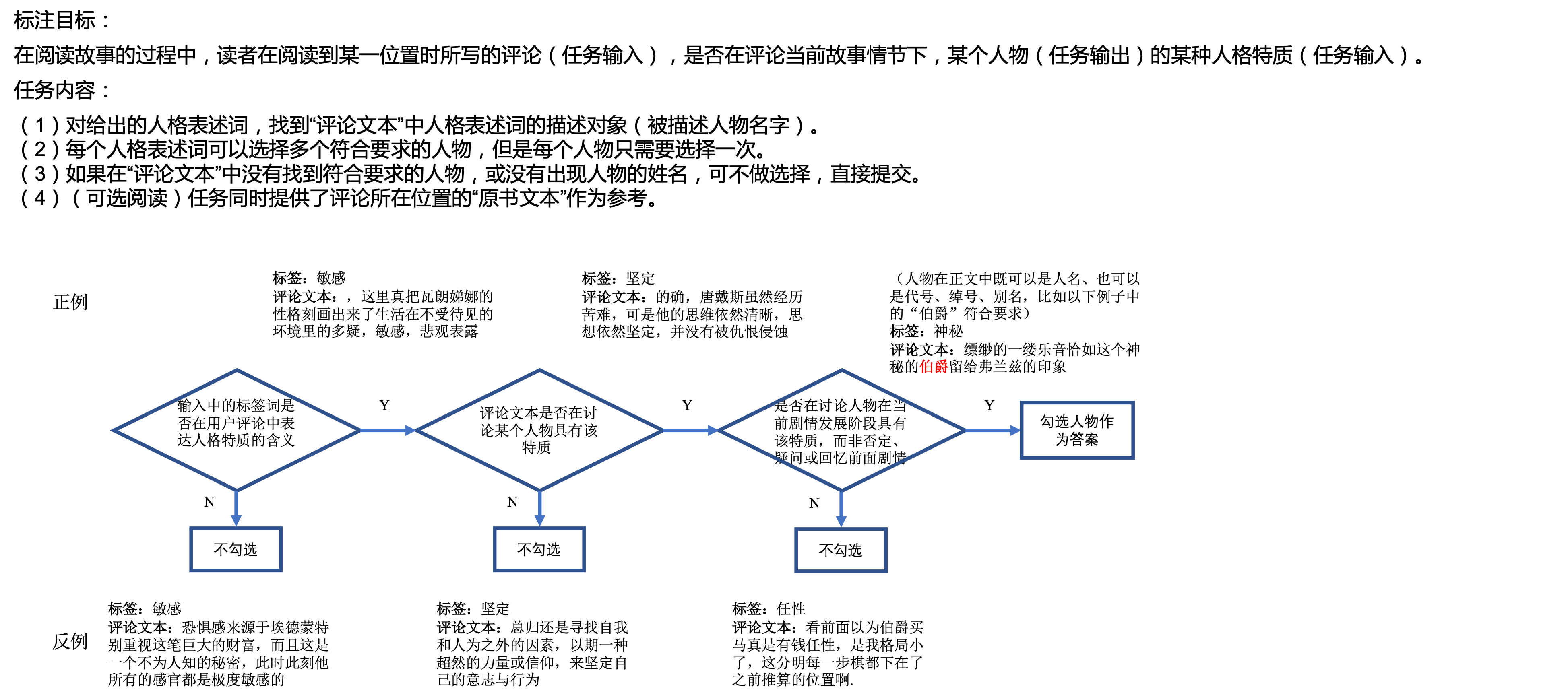}
\includegraphics[width=\textwidth]{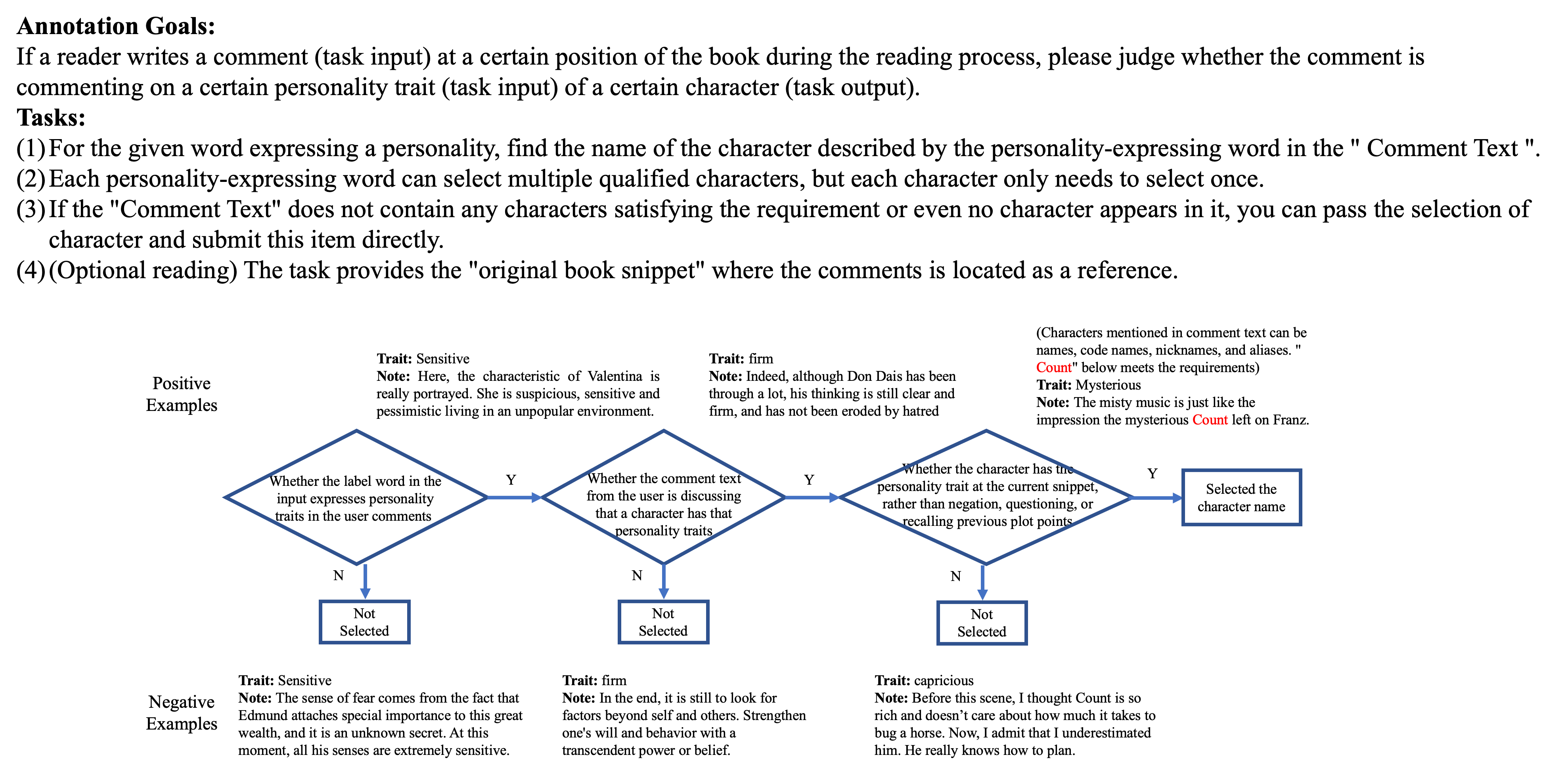}
\caption{\small{Our annotation guidelines. Top: the original Chinese guidelines. Bottom: the English translation.}}
\centering
\label{fig:guideline}
\end{figure*}

\begin{figure*}
\setlength{\belowcaptionskip}{-.5\baselineskip}
\centering
\includegraphics[width=\textwidth]{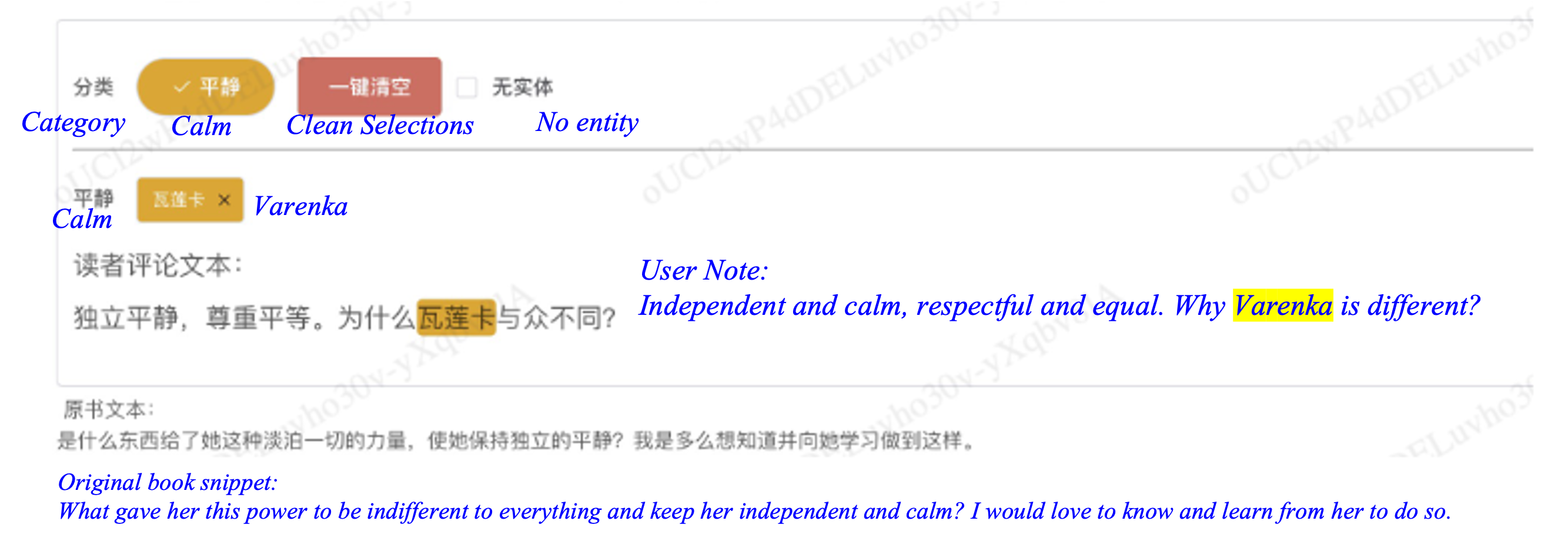}
\caption{\small{Our annotation interface (with English translations in \textcolor{blue}{\emph{blue words}}).}}
\centering
\label{fig:interface}
\end{figure*}

We show our guidelines in Figure~\ref{fig:guideline}; and the annotation interface with translations in Figure~\ref{fig:interface}.

\section{Notes that Are Difficult or Ambiguous to Label}
\label{app:difficult_cases}
\begin{CJK*}{UTF8}{gbsn}
Table~\ref{tab:ambiguous_cases} in Appendix~\ref{app:difficult_cases} gives some difficult examples that created disagreements.
There are two majors sources of difficulties. 
First, the trait word has multiple meanings in Chinese and the usage does not represent the sense of the trait.
In the first example, ``可怕的敌人\,(frightening enemy)'' in Chinese usually means ``an very-strong enemy that is hard/impossible to beat'', \emph{i.e.}, a terrible enemy.
The enemy, here refers to the protagonist Dant\`{e}s, does not necessary has the \emph{frightening} personality.
Similarly, in the second example, the annotators have disagreement because some people believe in Chinese, ``非凡\,(extraordinary)'' can be used as a personality trait only when a person possesses exceptional characteristics.
While some annotators think the trait can also describe a person with exceptional abilities.

Second, a trait word is used to recall the general impression or history behavior of a character in an implicit way.
In the third example, the user wanted to expresses that Elizabeth used to be clear-headed but becomes a fool at the dance party. 
This recall of the general impression \emph{clear-headed} is not explicit, but can be inferred from the next sentence that this note is commenting on a snippet of the dance party.
Therefore the user aims to comment the \emph{foolish} trait on the snippet instead of \emph{clear-headed}.

\begin{table}[h]
\small
\begin{center}
\begin{tabular}{||p{2.8in}||} 
 \hline
  {\textbf{Target Trait:} 可怕\,(\emph{frightening})} \\
  \textbf{Note Text (Chinese):} \textit{反目的朋友才是最\textcolor{blue}{可怕}的敌人，因为最了解你的是朋友，知道你短板最多的也是朋友，螳螂是唐格拉尔，黄雀是唐代斯}\\
  \textbf{Note Text (Translated):} \textit{A friend who turns against you is the most \textcolor{red}{terrible} enemy, because the friend who knows you best is the friend, and the friend who knows your most shortcomings is also the friend. The praying mantis is Tanglar, and the oriole is Dantes.}\\
 \hline\hline
  {\textbf{Target Trait:} 非凡\,(\emph{extraordinary})} \\
  \textbf{Note Text (Chinese):} \textit{维尔福的政治头脑在这一刻发挥了它最大的用处，任何时候都是极端的利己主义者，相对应也一定有\textcolor{blue}{非凡的时局判断力}，才能在哪儿都保全的了自己}\\
  \textbf{Note Text (Translated):} \textit{Villefort's political mind is at its best use at this moment, he is always an extreme egoist, correspondingly, he must have \textcolor{red}{extraordinary judgment of the situation}, in order to be able to protect himself everywhere.}\\
 \hline\hline
 {\textbf{Target Trait}: 清醒\,(\emph{clear-headed})} \\
 \textbf{Note Text (Chinese):} \textit{对于别人的事情，伊丽莎白却又\textcolor{blue}{清醒}得很，对于自己的事情却变成一个小傻瓜。这一次舞会在众人面前的出丑，或许成为了彬格莱先生后来的一次不了了之的决定性因素"} \\ 
 \textbf{Note Text (Translated):} \textit{For others' matters, Elizabeth is quite \textcolor{red}{clear-headed}, but for her own matters she becomes a fool. This time the embarrassment in front of everyone at the dance party, may have become a decisive factor for Mr. Bingley's later decision to forget about it.}\\
 \hline
\end{tabular}
\end{center}
\caption{Example of a human mistake.}
\label{tab:ambiguous_cases}
\end{table}
\end{CJK*}

\section{Full Book List}
\label{app:book_list}
Table~\ref{tab:detailed_books} shows the detailed information of each book included in our \datasetnamens.

\section{Visualization}
\label{app:visualization}
\begin{figure*}
\centering
\subfigure[\scriptsize{Dant\`{e}s (\emph{i.e.}, Count Monte Cristo)}]{\includegraphics[width=0.32\textwidth]
{figures/The_Count_of_Monte_Cristo_by_Alexandre_Dumas_Pere_Dantes.png}}\
\subfigure[\scriptsize{Albert (\emph{The Count of Monte Cristo})}]{\includegraphics[width=0.32\textwidth]
{figures/The_Count_of_Monte_Cristo_by_Alexandre_Dumas_Pere_Albert.png}}\
\subfigure[\scriptsize{Mr. Darcy (\emph{Pride and Prejudice})}]{\includegraphics[width=0.32\textwidth]{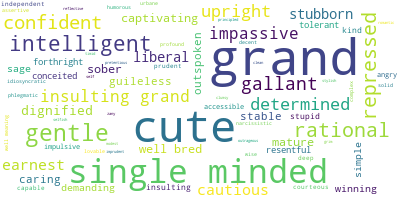}}
\subfigure[\scriptsize{Elizabeth (\emph{Pride and Prejudice})}]{\includegraphics[width=0.32\textwidth]{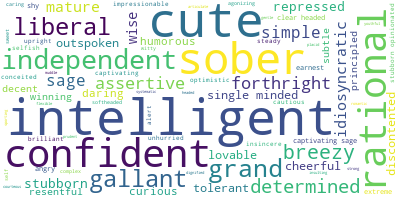}}
\subfigure[\scriptsize{Quasimodo (\emph{Notre-Dame de Paris})}]{\includegraphics[width=0.32\textwidth]{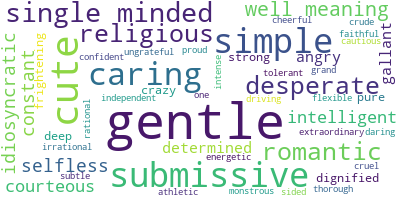}}
\subfigure[\scriptsize{Claude Froll (\emph{Notre-Dame de Paris})}]{\includegraphics[width=0.32\textwidth]{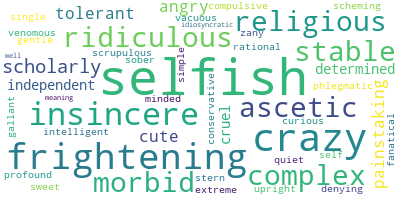}}
\vspace{-0.1in}
\caption{\small{Word clouds for the characters.}}
\centering
\label{fig:word_clouds_more}
\end{figure*}

\paragraph{Trait Clouds}
Figure~\ref{fig:word_clouds_more} include more word clouds for different characters.

\begin{figure*}
  \begin{minipage}[b]{0.52\textwidth}
\tiny
\centering
    \begin{tabular}{p{1.35in}ccc} 
        \toprule
        \bf Book Name &\bf \#Characters &\bf \#Instances & \bf \#Sentences\\ 
        \midrule
        \multicolumn{4}{c}{\emph{\underline{Training Books}}}\\
        \emph{Of Human Bondage} & 18&6539&16542\\
\emph{Pride and Prejudice} & 21&8632&5954\\
\emph{Madame Bovary} & 10&5440&6952\\
\emph{Anna Karenina} & 14&8204&20898\\
\emph{Anne Of Green Gables} & 8&3755&6834\\
\emph{Little Women} & 7&2726&9409\\
\emph{War and Peace} & 21&3448&31784\\
\emph{Don Quixote} & 3&767&9384\\
\emph{Wuthering Heights} & 13&2471&6753\\
\emph{Jane Eyre} & 10&1048&9692\\
\emph{Twenty Thousand Leagues under the Sea} & 4&471&6614\\
\emph{Jude the Obscure} & 3&174&9191\\
\emph{The Sorrows of Young Werther} & 1&74&2400\\
\emph{Father Goriot} & 5&198&6678\\
\emph{Uncle Tom's Cabin} & 5&138&10122\\
\emph{Vanity Fair} & 5&171&13125\\
\emph{Oliver Twist} & 5&178&9166\\
\multicolumn{4}{c}{\emph{\underline{Development Books}}}\\
\emph{The Red and the Black} & 6&721&11061\\
\emph{The Count of Monte Cristo} & 27&2488&26437\\
\emph{The Adventures of Tom Sawyer Complete} & 2&115&4913\\
\emph{David Copperfield} & 15&312&19195\\
\emph{The Gadfly} & 1&76&6875\\
\emph{A Tale of Two Cities} & 3&33&7757\\
\multicolumn{4}{c}{\emph{\underline{Testing Books}}}\\
\emph{Crime and Punishment} & 18&1498&14347\\
\emph{The Brothers Karamazov} & 12&638&24101\\
\emph{Les Miserables} & 14&557&35139\\
\emph{Eugenie Grandet} & 4&217&3797\\
\emph{Tess of the d'Urbervilles} & 3&162&8074\\
\emph{Notre-Dame de Paris} & 8&206&11278\\
\emph{The Call of the Wild} & 1&42&1696\\
\emph{The Idiot} & 9&215&16072\\
\emph{Moby Dick; or The Whale} & 2&51&9911\\
\emph{Resurrection} & 2&38&9760\\
        \bottomrule
    \end{tabular}
    \captionof{table}{\small{Detailed information of books included in our \datasetnamens.}}    \label{tab:detailed_books}
\end{minipage}
\hspace{0.06in}
  \begin{minipage}[b]{0.46\textwidth}
\small
\centering
\subfigure[\scriptsize{Dant\`{e}s and Albert from \emph{The Count of Monte Cristo}}]{\includegraphics[width=0.9\textwidth]
{figures/The_Count_of_Monte_Cristo_by_Alexandre_Dumas_Pere.png}\label{senti_a}}\vspace{-0.3cm}
\subfigure[\scriptsize{Frollo and Quasimodo from \emph{Notre-Dame de Paris}}]{\includegraphics[width=0.9\textwidth]{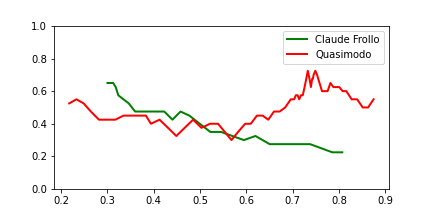}\label{senti_c}}
\caption{\small{Plots of sentiments of characters' traits along time. }}
\label{fig:sentiments_full}
\end{minipage}
\end{figure*}

\paragraph{Sentiment Plots}
Our trait vocabulary contains in total {818} traits with polarity annotations. 
Specifically, there are 234 positive traits, 292 neutral traits and 292 negative traits.
Figure~\ref{fig:sentiments_full} visualizes readers' sentiments towards four popular characters through the lens of traits.
We map the labeled traits to their sentiments, \emph{i.e.}, positive or negative, and then plot the sentiment along time. Here the x-axis is the position of an note with trait label, normalized by the lengths of the books.
The curves are smoothed within a window of 50 for \emph{The Count of Monte Cristo} and 20 for \emph{Notre-Dame de Paris}, depending on the sparsity of the samples.

The trends well align with the common feelings of readers during the reading process.
For example, the character \emph{Albert} is in general a brave and decent person. Most readers liked his personality until he recklessly challenged \emph{Dant\`{e}s} for a duel. Then the character's reputation is saved after he found out that his father framed many people including \emph{Dant\`{e}s} and decided to give up the duel and live off his father's ill-gotten gains.
One the other hand, \emph{Claude Frollo} received monotone decreased rates, who appeared first as a pious and highly knowledgeable man then turned to be evil and morbid because of his obsessive for \emph{Esmeralda}.

\begin{table}[t!]
\setlength{\belowcaptionskip}{-.5\baselineskip}
\setlength{\abovecaptionskip}{0.5\baselineskip}
    \small
    \centering
    \renewcommand{\arraystretch}{1} 
        \begin{tabular}{lcc} 
        \toprule
        \bf System & {\bf Slump}  & {\bf All}  \\ 
        \midrule
        BERT (no-hist)  & 35.98 & 40.33 \\
        \quad + unsup  & 56.25 &  44.58 \\
        Char-Hist-Longformer & 46.53 &45.75 \\
        \bottomrule
    \end{tabular}
    \caption{\small{Accuracy on the slump of Figure~\ref{plot_a} for the character \emph{Albert} (144 instances) versus on all (424) of the \emph{Albert} instances.}}
    \label{tab:slump_performance}
\end{table}
\paragraph{A Case Study}
We assessed the model performance on the points where people's ratings of \emph{Albert} have dramatic fluctuations (around $x$=0.8).
Specifically, we compared three models: the baseline BERT model without any history, the BERT model enhanced with our unsupervised objective, and the Char-Hist Longformer, which can leverage longer historical information.
The results are shown in Table~\ref{tab:slump_performance}.

Our findings revealed that both the enhanced models—BERT with the unsupervised objective and Char-Hist Longformer—achieved a similar level of improvement over the BERT baseline when considering the entire evaluation set of \emph{Albert}. These results align with our experimental observations from the comprehensive evaluation data.
However, it is noteworthy that the model incorporating the unsupervised objective exhibited a significantly greater enhancement at the slump of the curve.
As mentioned earlier in this section, the author explicitly portrayed Albert's reckless personality through his actions and dialogues in this particular case. Even without prior knowledge of the events leading up to this point, humans can intuitively grasp Albert's personality traits. Our unsupervised task aims to capture the correlation between personality and the external expressions manifested within the narrative. This is why it proves to be more effective in this specific case.

\section{Examples of Cases that Require History Information}
\label{app:examples_history}

The cases where history information is necessary to solve can be roughly categorized into three types according to our human study in Section~\ref{ssec:exp_human_study}.
We include examples for each type in Table~\ref{tab:cases_need_history}.

\begin{table*}[h]
\small
\begin{center}
\begin{tabular}{||p{5.8in}||} 
 \hline
 \textbf{Category: }\emph{(1) multiple possible answers given the snippet without history}\\
{\textbf{Target Character:} \emph{Dant\`{e}s}} \quad
  {\textbf{Groundtruth Trait:} \emph{simple}} \\
  \textbf{Distractors:} \emph{insincere, dirty, impressionable, loquacious} \\
  \textbf{Snippet:} \textit{i am the abbe faria, and have been imprisoned as you know in this chateau d ’ if since the year 1811 ; previously to which i had been confined for three years in the fortress of fenestrelle. in the year 1811 i was transferred to piedmont in france. it was at this period i learned that the destiny which seemed subservient to every wish formed by napoleon, had bestowed on him a son, named king of rome even in his cradle. i was very far then from expecting the change you have just informed me of ; namely, that four years afterwards, this colossus of power would be overthrown. then who reigns in france at this moment — napoleon ii.? ” “ no, louis xviii. ” ...
  \textcolor{red}{dantes ’ whole attention was riveted on a man who could thus forget his own misfortunes while occupying himself with the destinies of others.}
  “ yes, yes, ” continued he, “ ’ twill be the same as it was in england. after charles i., cromwell ; after cromwell, charles ii., and then james ii., and then some son - in - law or relation, some prince of orange, a stadtholder who becomes a king. then new concessions to the people, then a constitution, then liberty. ah, my friend! ” said the abbe, turning towards dantes, and surveying him with the kindling gaze of a prophet, “ you are young, you will see all this come to pass. ...”}\\
 \hline\hline
  \textbf{Category: }\emph{(2) plot cannot be understood without history}\\
  {\textbf{Target Character:} \emph{The elder}} \quad
  \textbf{Groundtruth Trait:} \emph{intelligent} \\
  \textbf{Distractors:} \emph{confident, breezy, single-minded, decadent} \\
  \textbf{Snippet:} \textit{
the servant soon returned. the decanter and the glass were completely empty. noirtier made a sign that he wished to speak. “ why are the glass and decanter empty? ” asked he ; “ valentine said she only drank half the glassful. ” the translation of this new question occupied another five minutes. “ i do not know, ” said the servant, “ but the housemaid is in mademoiselle valentine ’ s room : perhaps she has emptied them. ” “ ask her, ” said morrel, translating noirtier ’ s thought this time by his look. the servant went out, but returned almost immediately. “ mademoiselle valentine passed through the room to go to madame de villefort ’ s, ” said he ; “ and in passing, as she was thirsty, she drank what remained in the glass ; as for the decanter, master edward had emptied that to make a pond for his ducks. ” noirtier raised his eyes to heaven, as a gambler does who stakes his all on one stroke.
\textcolor{red}{from that moment the old man ’ s eyes were fixed on the door, and did not quit it.}
it was indeed madame danglars and her daughter whom valentine had seen ; they had been ushered into madame de villefort ’ s room, who had said she would receive them there. that is why valentine passed through her room, which was on a level with valentine ’ s, and only separated from it by edward ’ s. the two ladies entered the drawing - room with that sort of official stiffness which preludes a formal communication. among worldly people manner is contagious. madame de villefort received them with equal solemnity. valentine entered at this moment, and the formalities were resumed. ... 
}\\
 \hline
 \hline
  \textbf{Category: }\emph{(3) The current snippet can be associated to some previous plot where the character demonstrates a personality trait}\\
  {\textbf{Target Character:} \emph{Esmeralda}} \quad
  \textbf{Groundtruth Trait:} \emph{simple} \\
  \textbf{Distractors:} \emph{rational, mature, emotional, egocentric} \\
  \textbf{Snippet:} \textit{
is she to be hung yonder? " " fool! t'is here that she is to make her apology in her shift! the good god is going to cough latin in her face! that is always done here, at midday. if'tis the gallows that you wish, go to the greve. " " i will go there, afterwards. " " tell me, la boucanbry? is it true that she has refused a confessor? " " it appears so, la bechaigne. " " you see what a pagan she is! " "'tis the custom, monsieur. the bailiff of the courts is bound to deliver the malefactor ready judged for execution if he be a layman, to the provost of paris ; if a clerk, to the official of the bishopric. " " thank you, sir. " " oh, god! "
\textcolor{red}{said fleur - de - lys, " the poor creature! " this thought filled with sadness the glance which she cast upon the populace. the captain, much more occupied with her than with that pack of the rabble, was amorously rumpling her girdle behind. she turned round, entreating and smiling. " please let me alone, phoebus! if my mother were to return, she would see your hand! "}
at that moment, midday rang slowly out from the clock of notre - dame. a murmur of satisfaction broke out in the crowd. the last vibration of the twelfth stroke had hardly died away when all heads surged like the waves beneath a squall, and an immense shout went up from the pavement, the windows, and the roofs, " there she is! " fleur - de - lys pressed her hands to her eyes, that she might not see. " charming girl, " said phoebus, " do you wish to withdraw? " " no, " she replied ... 
}\\
 \hline
\end{tabular}
\end{center}
\caption{\small{Example of cases that require history information to solve. The \textcolor{red}{red} texts are the underlined text of the notes that used to construct the labeled instance. In the first example, according to the snippet, both \emph{simple} and \emph{impressionable} are possible traits to explain the character's behavior. Only from the history that \emph{Dantes} is a brave and determined person, we can select \emph{simple} as the correct answer.} In the second example, only when the readers know that \emph{Noirtier} (\emph{The elder}) aims to help Valentine get immunity from the poisoned juicy, they can understand the character's wisdom. In the third example, \emph{Esmeralda} is not present. However, the scene of love between Phoebus and Fleur-de-Lys is quite similar to her story with Phoebus, illustrating that she was easily deceived by the man.}
\label{tab:cases_need_history}
\end{table*}

\end{document}